\documentclass[a4paper]{article}
\usepackage{spconf}
\usepackage{multirow}
\usepackage{float}
\usepackage{hyperref}
\usepackage{graphicx}
\usepackage{tikz}
\usetikzlibrary{positioning}
\usepackage[compact]{titlesec}

\hypersetup{
colorlinks=true,
urlcolor=cyan,
}

\title{A comparison of methods for OOV-word recognition on a new public dataset}
\name{Rudolf A. Braun, Srikanth Madikeri, Petr Motlicek\thanks{Thanks to Phil Garner for the helpful discussions}}
\address{Idiap Research Institute, Martigny, Switzerland}

\begin{document}
\ninept
 
\maketitle

\begin{abstract}
A common problem for automatic speech recognition systems is how to recognize words that they did not see during training. Currently there is no established method of evaluating different techniques for tackling this problem.\\
We propose using the CommonVoice dataset to create test sets for multiple languages which have a high out-of-vocabulary (OOV) ratio relative to a training set and release a new tool for calculating relevant performance metrics. We then evaluate, within the context of a hybrid ASR system, how much better subword models are at recognizing OOVs, and how much benefit one can get from incorporating OOV-word information into an existing system by modifying WFSTs. Additionally, we propose a new method for modifying a subword-based language model so as to better recognize OOV-words. We showcase very large improvements in OOV-word recognition and make both the data and code available.
\end{abstract}


\textbf{DOI: 10.1109/ICASSP39728.2021.9415124}

\section{Introduction}

All languages are constantly evolving and therefore all ASR systems suffer from failing to detect words that were not in their training set (out-of-vocabulary, OOV, words). We focus on weighted finite-state transducer (WFST) based ASR systems with distinct acoustic and language models~\cite{Mohri2008}. In these systems both the language model and lexicon are fixed and encoded as a WFST, this means words that were not part of these systems at training time are impossible to recognize. This has lead to various approaches to modify the WFSTs so that the ASR system can recognize words it had previously no knowledge of~\cite{aleksic_ctxinfo,aleksic_impr_rec,novak_dynamic,allauzen_rapidadd, speech2003, bulusheva}. A complication is that typically the lexicon and language model WFSTs will be composed together to create a static decoding graph that can be used repeatedly during decoding. This is a problem because, depending on the use-case, it means we don't have access to the lexicon WFST (L) or the language model WFST (G), and must try and alter the full decoding graph, the HCLG, which is harder.

One workaround is to use a subword-based model, as they can theoretically create any word by outputting a sequence of shorter subword tokens~\cite{thomas_ibm_oov,He_subword_oov,bisani_openvoc}.
Another approach is for the language model to contain a $[unk]$ (unknown) token, which has as the pronunciation a phone LM trained on a lexicon of words with low counts, and then to try recover a word from the recognized phone sequence aligned with the $[unk]$ token~\cite{bazzi_oov, alumae_est}.

Doing graph composition on a client's device can be difficult as it can take a significant amount of time and memory to perform. Therefore it is usually preferred to deploy ASR systems with an already composed decoding graph. If one is willing to redo composition but does not want to retrain the language model modifying the L and G directly is an option. Alternatively, one can avoid having to create the static decoding graph by doing on-the-fly composition, also known as dynamic composition, which is done at runtime~\cite{dynamic-riley, dynamic-idiap, riley_preinit,fb_dynamic}. Keeping the G, for example, separate makes it easier to bias the model towards certain words or add new ones to it~\cite{aleksic_ctxinfo, aleksic_impr_rec, novak_dynamic}. However, this approach causes a decrease in decoding speed.

Finally, one can try and modify the static decoding graph (HCLG)~\cite{allauzen_rapidadd, speech2003, bulusheva}. Because of composition and optimization (e.g. determinisation, minimisation, weight-pushing) the initially separated knowledge sources (the lexicon, language model, etc.) are now entangled, making it harder to modify or add new words and pronunciations than when working with the separated L and G.

Many existing papers focusing on OOV recognition used private datasets, which makes results not comparable~\cite{aleksic_ctxinfo, allauzen_rapidadd, thomas_ibm_oov, bazzi_oov}. Or to create OOVs they keep the top ten thousand (or some other number that is significantly smaller than a real ASR system would use) in the vocabulary and use the rest as OOV words~\cite{novak_dynamic, allauzen_rapidadd, bisani_openvoc, thomas_ibm_oov}. This evaluation method is problematic because it would overestimate the benefit of using subword-based models as relatively frequent words are not included in the top ten-thousand but the various inflections of them will be seen often during LM training by the subword-based model. This will make it artificially easy to then recover the OOV-word as the subword sequence needed will have a relatively high probability. For the same reason (these artificial OOV-words actually being common when considering inflections) grapheme2phoneme tools will return more accurate pronunciations than would happen with realistic OOV-words.

Therefore, we create reproducible datasets for English and German using CommonVoice\cite{commonvoice} where the test set has a large number of realistic OOVs. We release a new tool for calculating error rate metrics, and propose a new metric called ``OOV-CER" for measuring OOV-word recognition performance independent of the performance on in-vocabulary words. Using this setup we compare word to subword-based models, check how well OOV recognition works when using a phone LM as the pronunciation for $[unk]$, and compare how effective modifying the L, G and HCLG is. Finally, we propose a new method for modifying the G of a subword-based model to improve performance.\\
The data and relevant code to modify WFSTs (discussed later) can be found here: \href{github.com/idiap/icassp-oov-recognition}{github.com/idiap/icassp-oov-recognition}.

\section{Dataset}

The goal is to create a test dataset with a high amount of realistic OOV-words. The approach we use is to have a large vocabulary and then choose utterances from the CommonVoice\cite{commonvoice} dataset that contain at least one OOV-word to create the test dataset. The training set is created from the remainder, while excluding those utterances that would lead to a speaker overlap between train and test. For English we used the Librispeech\cite{librispeech} lexicon as the vocabulary, for German we created one by taking the top two hundred thousand words from a text corpus (Europarl). By using large vocabularies gotten from large corpora we ensure that any OOVs will be realistic.\\
The training and test set size is 280 / 250 and 2.5 / 3 hours for English / German respectively. The OOV ratio is 12.2 / 13.6\%.
The distribution of the OOV-words is very flat. The English ones tend to be modern words, the top three are ``firefox'', ``website'' and ``nudism''. This is because the Librispeech corpus is based on old books, so the vocabulary is old-fashioned. The German OOV-words tend to be compounds words. The English task is harder as the test set text is not only a different domain but also from a different time-period than the vocabulary and text corpus used to train the LM.\\

\section{Metrics}

We measure the standard WER (word error rate) and CER (character error rate). Character error rate is a useful measure because if a word has one character wrong that should be a less significant error than if most are incorrect. Additionally, it is useful to know how well OOV-words are recognized independent of performance on in-vocabulary words because OOV-words are more important than for example stop words (``the'', ``a'', ``and'' etc.). This could be done by measuring OOV recall (how many times a OOV-word in the reference is predicted) but this, like WER, treats one or five character mistakes equally.
Therefore we developed a new tool for calculating error metrics and propose a new metric called `OOV-CER'. The tool is called \texttt{texterrors} and is available at \href{https://github.com/RuABraun/texterrors}{github.com/RuABraun/texterrors}.\\
It does character aware alignment of the reference and hypothesis by incorporating the edit distance between words into the substitution cost. The OOV-CER is calculated by getting the index of the OOV-word in the reference, using it to index into the aligned hypothesis and then calculating the edit distance between that word and the reference word. To take into account that a model could output the reference as two separate words, words in the aligned hypothesis that neighbor the index (obtained from where the OOV-word is in the aligned reference) and are aligned with nothing (are insertions) will be pre- or appended to the word in the index.\\
As an example: The reference is "words in sentence", the hypothesis is "words in sent tense" and the word "sentence" is the OOV-word and is aligned to "sent". To calculate the OOV-CER we first join "sent" and "tense", as the latter is an insertion and aligned next to the OOV-word, and then calculate the CER between "sentence" and the joined word.\\

We don't bother measuring OOV precision as a decrease in performance will already be reflected by an increase in WER/CER. As OOV-words are more important than most in-vocabulary words if the OOV-CER goes down while the WER stays the same after applying some modification to the model, we consider the model as improved.\\

\section{Model biasing mechanisms}

A very common use-case is to have some prior knowledge about likely OOV-words, and to want to adjust the model so as to recognize them. In this section, we first review three approaches and introduce a new one. When we mention using a list of OOV-words, we mean a list that has been extracted from the test set relative to our model vocabulary. This is therefore the best case scenario as we know all OOV-words that our model will be asked to recognize. The $[unk]$ symbol is a token that represents an unknown word, $jnk$ is its default pronunciation.

\subsection{UNK with non-jnk pronunciation}

\begin{figure}[h]
    \centering
    \includegraphics[scale=0.4]{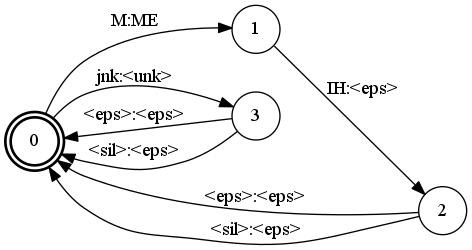}
    \caption{Simple example of a lexicon WFST (the L).}
    \label{fig:lfst}
\end{figure}

This method does not actually require any knowledge of possible OOV-words in advance. Rather than having $jnk$ be the pronunciation of the $[unk]$ token, one can replace it with a phone LM trained on the phones from a lexicon of (possible OOV-) words. The LM is inserted in WFST form. Our implementation uses kaldi's \texttt{utils/lang/make\_unk\_lm.sh}. This allows for an almost arbitrary phone sequence to be recognized.\\In figure \ref{fig:lfst} one can see a simple L. If we wanted to insert just one pronunciation for $[unk]$ we would delete the existing arc from state 0 to 3, then add an arc for each phone in the pronunciation starting from state 0 and ending at state 3. One of these would have $[unk]$ as the output label. To add a phone LM we take an existing WFST over phones P, and connect state 0 in the L to the start state of P with $[unk]$ as the output label, then connect all final states of P to state 3. The connecting arcs will have input disambiguation symbols to ensure the L is still determinisable.\\
After decoding one then aligns the best-path output lattice to find which phones match to $[unk]$, runs phoneme2grapheme (trained separately), rescores the alternatives with a character LM and gets the best path to get the recovered word. When the training data for the phone LM comes from the lexicon of OOV-words we call this method 'biased unk lm'. To simulate the case when we don't know what words are OOV we get phones from a lexicon of words with low counts (relative to the text corpus used to train the LM) and call it `unk lm'.
    
\subsection{Replacing UNK in L and G}

This approach assumes one has access to the L and G WFSTs. Using a lexicon of all OOV-words, we add the words and corresponding pronunciations into the L. This is easy to do as the L is unoptimized and we can just add the pronunciations as a sequence of arcs with one of them having the word as the output label. It assumes the new words do not contain any new phones. Then we iterate over the states of the G and replace all arcs with $[unk]$ with multiple arcs keeping the same start and end state, each with one of the OOV-words we want to add as the input and output label. Each arc inherits the $[unk]$ weight plus a penalty of 2.3 (equivalent to multiplying the probability by 0.1). The penalty is because $[unk]$ has a relatively high probability, and we empirically found this to help. This method is called `mod L,G'. 

\subsection{Replacing UNK in HCLG}

To replace the \texttt{jnk:[unk]} arcs in the HCLG we need an HCL, as the input labels of the HCLG are transition-ids and the states represent different HMM states. We can create an HCL from the lexicon of OOV-words and then do the replacement. For the sake of simplicity our method requires that the HMM topology only has one state. Doing the replace operation makes an additional assumption which constrains the sort of models we can use: By default our models use biphone context dependency, now imagine we inserted the HCL of a word who's pronunciation started with some phone $p$, the issue is that the input label associated with $p$ should be different depending on what arc came before (i.e. what phone came before) the one we are replacing in the HCLG. But we can't know that at the time of the HCL creation. We get around this problem by using a monophone model. While techniques exist to modify the HCLG of context dependent models~\cite{allauzen_rapidadd}\cite{bulusheva} they are quite complex and we want to test whether using context dependency is even necessary. Due to our LM being trained with the \texttt{limit-unk-history} option of \textit{pocolm}, $[unk]$ can only appear at the end of an ngram, so we can just insert the HCL once, and point all arcs matching $[unk]$ to it. The outgoing arcs have the same probability for all histories, as there are no saved histories for $[unk]$. This means the HCLG barely changes in size after the operation. As in `mod L G' we add in a penalty of 2.3. This method is called `mod HCLG'.

\subsection{Modifying subword G}

Trying to modify a word-based model so as to incorporate prior knowledge and better recognize certain (possibly OOV) words is a common focus. However we are not aware of any efforts to try the same with a subword-based model. Since subword-based models can outperform word-based models when there are many OOVs (see section 6), we decided to try incorporate prior knowledge to improve performance even more. We do this by modifying the G (this assumes the G is available separately). We tokenize each OOV-word, and then check if that sequence of subwords exists in the $G$ starting from the backoff state. If it does, we lower the cost (cost because weights are the negative log of the probability) slightly, if it does not we add the necessary arcs with a low cost. The final arc goes to the unigram state of the last subword. This method is called `mod G'.

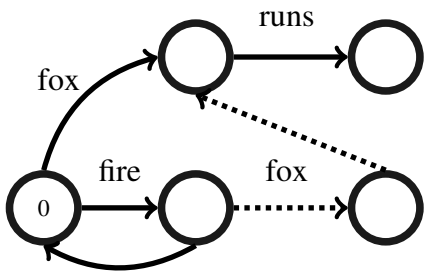
\begin{figure}[h]
    \centering
\begin{tikzpicture}[node/.style={circle, draw=black!90, fill=white!5, line width=1mm, minimum size=9mm, on grid},node distance=2cm]
\node[node](start){0};
\node[node](fire) at (2,0){};
\node[node](fox) at (2,2) {};
\node[node](runs) at (4.5,2) {};
\node[node](fire_fox) at (4.5,0){};

\node at (1.,0.5) {\large fire};
\node at (0.2,1.7) {\large fox};
\node at (3.2,2.5) {\large runs};
\node at (3.2,0.5) {\large fox};

\draw[->,line width=2pt](start.north) to[bend left] (fox.west);
\draw[->,line width=2pt](start.east) -- (fire.west);
\draw[->,line width=2pt](fire.south) to[bend left] (start.south);
\draw[->,line width=2pt](fox.east) -- (runs.west);
\draw[dotted,->,line width=2pt](fire.east) -- (fire_fox.west);
\draw[dotted,->,line width=2pt](fire_fox.north) to (fox.south);
\end{tikzpicture}
    \caption{Illustrative example of how `mod G' will modify the G by adding new arcs (dashed lines are new arcs) with low costs to increase the odds of recognising certain words. The 0 state is the start state.}
    \label{fig:mod_G}
\end{figure}

In figure \ref{fig:mod_G} a simplified G for illustrative purposes. The 0 state is the start state from which all unigram arcs start. By adding a new (represented by a dashed line) arc `fox' with a low cost (high probability) from the `fire' unigram state we lower the total cost of recognizing the subwords `fire' and `fox', thereby making it easier for the model to recognize the OOV-word `firefox'. The alternative, going from the unigram state `fire' back to state 0 along the backoff arc and then to the unigram state `fox', would result in a higher total cost for the subword sequence. We also add a back-off arc going to the unigram state `fox', rather than back to the start state, so that the language model knows that the previous subword was `fox' which improves performance. 

\section{Experimental setup}

For both languages for the word-based models we train a trigram language model using pocolm, and prune to 3.5 million ngrams. The subword based model uses a five gram pruned to the same number. We use BPE to choose the set of subword tokens and allow 5000 merges. The lexicon of the subword-based model is character based (this performed better than using g2p on the subword tokens). For English the LM training data is the Librispeech text corpus and the we use the 200k lexicon that is part of the corpus, we create pronunciations for OOV words using Phonetisauras\cite{novak}. For German we use the Europarl corpus, the vocabulary is the top 200k words, we used espeak-ng for creating the pronunciations.\\
For training the acoustic model and doing decoding we use kaldi\cite{kaldi}. We follow the standard procedure of getting alignments via HMM-GMM training and then training a TDNNF\cite{tdnnf} model with LF-MMI\cite{LFMMI} and ivectors. We use biphone context dependency unless indicated otherwise.

\section{Results \& Discussion}

\subsection{No prior knowledge}

The first case we consider is when no knowledge about potential OOV-words is available. We want to test the assumption that subword-based models do better than word-based, and how well word recovery performs when using the 'unk lm' method. As mentioned previously when using the 'unk lm' method we train once on a lexicon of words with low counts, and once on the lexicon of OOV-words, the latter is 'biased unk lm'. By comparing the two we can test how important it is for the phone LM to be trained on phone sequences that equal the ones seen at test time. The results can be seen in table \ref{table:noprior}.

\begin{table}[H]
\centering
\begin{tabular}{|l|c|c|c|c|}
\hline
                         &      & WER  & CER  & OOV-CER \\ \hline
\multirow{4}{*}{English} & word & 36.3 & 19.7 & 54.1    \\ \cline{2-5} 
                         & word, unk lm & 35.9 & 18.6 & 51.8  \\ \cline{2-5} 
                         & word, biased unk lm & 35.4 & 18.7 & 52.0  \\ \cline{2-5} 
                         & BPE  & 37.2 & 19.1 & 52.1    \\ \hline\hline
\multirow{4}{*}{German}  & word & 29.9 & 10.2  & 44.4     \\ \cline{2-5} 
                         & word, unk lm & 26.9 & 9.2 & 37.2  \\ \cline{2-5} 
                         & word, biased unk lm & 25.6 & 8.8 & 34.7  \\ \cline{2-5}
                         & BPE  & 25.2 & 8.2 & 36.0  \\ \hline
\end{tabular}
\caption{Comparison of word- and subword-based models and OOV recovery using a phone LM when no prior information about OOV-words is known.}
\label{table:noprior}
\end{table}

Comparing word to subword-based models there is no improvement for English but a significant one for German. These results make sense as the types of OOV-words differ between the two languages. In German a lot of the OOV-words are compounds words, these words can be created by a sequence of subwords which themselves are valid words in the German language and are therefore more likely to be present in ngrams of the ngram language model. In general subword-based LMs benefit from the fact that, unless a character in a word is very unusual, every word in the training set will be used for training (in segmented form), whereas word LMs will convert all words not part of the vocabulary to $[unk]$.\\
In the English dataset the OOV-words tend to be completely novel. This means the subword LM is very unlikely to have seen the sequence of subwords, and since there is no natural way to split the OOV-words (because they are not compound words) it is likely that the subwords needed to create the OOV-word will be short (which makes it harder for the language model to make estimates, consider the extreme case of a word being split up into individual characters to understand why), and that no or few n-grams contain these subwords, leading to the language model assigning the subword sequence a low probability.\\
With the `unk lm' method one can see an insignificant benefit for English and a noticeable one for German. We decided to test whether the issue was the phone based lexicon for English, and therefore trained a model that used characters as pronunciations. This meant we did not need to do any sort of g2p to get pronunciations for words not in the librespeech lexicon, or do p2g when doing OOV recovery to convert a recognized phone sequence back to letters. We just need to find the characters aligned to $[unk]$ in the decoded lattice. We trained the char LM that is the pronunciation of $[unk]$ on the OOV-word character lexicon. Table \ref{table:unklmchar} shows the results.

\begin{table}[H]
\centering
\begin{tabular}{|l|c|c|c|c|}
\hline
                         &      & WER  & CER  & OOV-CER \\ \hline
\multirow{5}{*}{English} & word & 36.3 & 19.7 & 54.1    \\ \cline{2-5} 
                         & word, unk lm & 35.9 & 18.6 & 51.8  \\ \cline{2-5} 
                         & word, biased unk lm & 35.4 & 18.7 & 52.0  \\ \cline{2-5} 
                         & word char & 37.0 & 19.4 & 53.3    \\ \cline{2-5} 
                         & word char unk lm & 36.0 & 18.8 & 50.4    \\ \hline
\end{tabular}
\caption{Comparing OOV recovery with a phone LM to using a model with a character based lexicon, where recovering the word is trivial}
\label{table:unklmchar}
\end{table}

The character based model doing OOV recovery does slightly better at recognizing OOV-words, but the WER is still close enough to the phone based baseline model that it is questionable whether the effort is worth it as this is the best case performance since the character LM (used as pronunciation for $[unk]$) was trained on the OOV-word character lexicon.
These results show that without having some prior knowledge about the OOV-words the model will encounter, it is very difficult for a hybrid based ASR system to deal with them. In languages with a significant amount of compound words one can use the just described methods to mitigate the amount of errors caused by OOV-words, but the improvement is moderate.

\subsection{With prior knowledge}

It is a very common use-case to know that certain OOV-words will need to be recognized by a model. We compare three different scenarios: When we have access to the L and G and are willing to redo composition ('mod L,G'), when we don't want to redo composition and therefore modify the HCLG and are willing to accept the constraint of using a monophone model ('mod HCLG'), and when we have a subword-based model and have access to the G and will do composition again ('mod G'). 
In each case we assume we have a list of OOV-words that we know the model will need to recognize, see section 4 for details on how to incorporate that information.
The results are in table \ref{table:modHCLG}.

\begin{table}[H]
\centering
\begin{tabular}{|l|c|c|c|c|}
\hline
                         &      & WER  & CER  & OOV-CER \\ \hline
\multirow{6}{*}{English} & word & 36.3 & 19.7 & 54.1    \\ \cline{2-5} 
                         & word mod L,G  & 24.3 & 13.8 & 16.1    \\ \cline{2-5}
                         & word mono & 36.8 & 19.2 & 53.2 \\ \cline{2-5}
                         & word mono mod HCLG  & \textbf{23.6} & \textbf{13.0} & \textbf{15.2} \\ \cline{2-5}
                         & BPE  & 37.2 & 19.1 & 52.1    \\ \cline{2-5}
                         & BPE mod G & 29.4 & 15.8 & 33.4 \\ \hline\hline
\multirow{6}{*}{German}  & word & 29.9 & 10.2  & 44.4     \\ \cline{2-5} 
                         & word mod L,G & 12.0 & \textbf{4.9} & 4.7  \\ \cline{2-5}
                         & word mono & 30.1 & 10.4 & 39.7 \\ \cline{2-5}
                         & word mod HCLG & \textbf{11.8} & 5.1 & \textbf{4.5}\\ \cline{2-5}
                         & BPE  & 25.2 & 8.2 & 36.0  \\ \cline{2-5}
                         & BPE mod G & 14.8 & 5.5 & 11.1 \\ \hline
\end{tabular}
\caption{Comparison of the baseline to 'mod L,G', a monophone baseline and 'mod HCLG', the BPE baseline and 'mod G' which modifies the subword-based model.}
\label{table:modHCLG}
\end{table}

All methods lead to a very large performance improvement on OOV-words. The fact that the monophone model is so competitive with the biphone baseline supports the modern trend of not using context dependent targets for the acoustic models\cite{wav2letter}\cite{espnet}, and suggests that these targets are more robust to out-of-domain data (as the OOV-CER is lower). The results also show that using the $[unk]$ probability is a legitimate approach for modeling OOV-words, which makes sense since words that will end up OOV tend to have certain characteristics like being nouns. Adding the penalty of 2.3 to the arcs of each added word improved performance by roughly 10\%.
While `mod G` improves the performance of the subword-based model significantly, the modifications for word-based models are better. We believe this is because a lot of OOV-words will be represented by several short subwords, and both their and the pronunciations of the OOV-word (as realized by connecting the pronunciations of the subwords) can be inaccurate, making it hard for the model to recognize the exact sequence of subwords needed to create the OOV-word.\\


\section{Conclusion}

We used CommonVoice to create shareable datasets for evaluating OOV-word recognition in English and German. Using a new tool \texttt{texterrors} we developed for calculating error metrics, we conducted experiments on OOV recognition performance across two languages in two different scenarios: Without and with prior knowledge. When no prior knowledge is available subword-based models and OOV-word recovery, with a phone LM for $[unk]$, improve results slightly. With prior knowledge we showed several methods to dramatically reduce the error rate on OOV-words. The best approach for dealing with a high OOV-ratio is to use a word-based, context independent model and a modified HCLG. We have shared the data and the code so that others can evaluate their own methods, compare to an existing baseline and build upon our results.

\bibliographystyle{IEEEbib}
\bibliography{bibliography}

\end{document}